\documentclass[twocolumn,english]{IEEEtran}
\usepackage[T1]{fontenc}
\usepackage{fancyhdr}
\usepackage{color}
\usepackage{array}
\usepackage{textcomp}
\usepackage{multirow}
\usepackage{amstext}
\usepackage{graphicx}
\usepackage{wasysym}

\makeatletter

\providecommand{\tabularnewline}{\\}

\usepackage{cite}

\usepackage{babel}
\begin{document}

\title{Energy-Efficient CMOS Memristive Synapses for Mixed-Signal Neuromorphic
System-on-a-Chip }

\author{{\normalsize{}Vishal Saxena, Xinyu Wu, and Kehan Zhu, }\emph{\normalsize{}Members
IEEE }\thanks{V. Saxena and X. Wu are with the Department of Electrical and Computer
Engineering, University of Idaho, Moscow ID 83844 (e-mail: vsaxena@uidaho.edu).
K. Zhu is now with Maxim Integrated. This work was supported in part
by NSF CAREER Grant EECS-1454411. }}
\maketitle
\begin{abstract}
Emerging non-volatile memory (NVM), or memristive, devices promise
energy-efficient realization of deep learning, when efficiently integrated
with mixed-signal integrated circuits on a CMOS substrate. Even though
several algorithmic challenges need to be addressed to turn the vision
of memristive Neuromorphic Systems-on-a-Chip (NeuSoCs) into reality,
issues at the device and circuit interface need immediate attention
from the community. In this work, we perform energy-estimation of
a NeuSoC system and predict the desirable circuit and device parameters
for energy-efficiency optimization. Also, CMOS synapse circuits based
on the concept of CMOS memristor emulator are presented as a system
prototyping methodology, while practical memristor devices are being
developed and integrated with general-purpose CMOS. The proposed mixed-signal
memristive synapse can be designed and fabricated using standard CMOS
technologies and open doors to interesting applications in cognitive
computing circuits.
\end{abstract}

\begin{IEEEkeywords}
CMOS Neurons, Memristors, Neuromorphic computing, Spiking Neural Networks
(SNNs), STDP, Synapses.
\end{IEEEkeywords}

\section{Introduction}

\begin{figure}[tb]
\begin{centering}
\includegraphics[width=1\columnwidth]{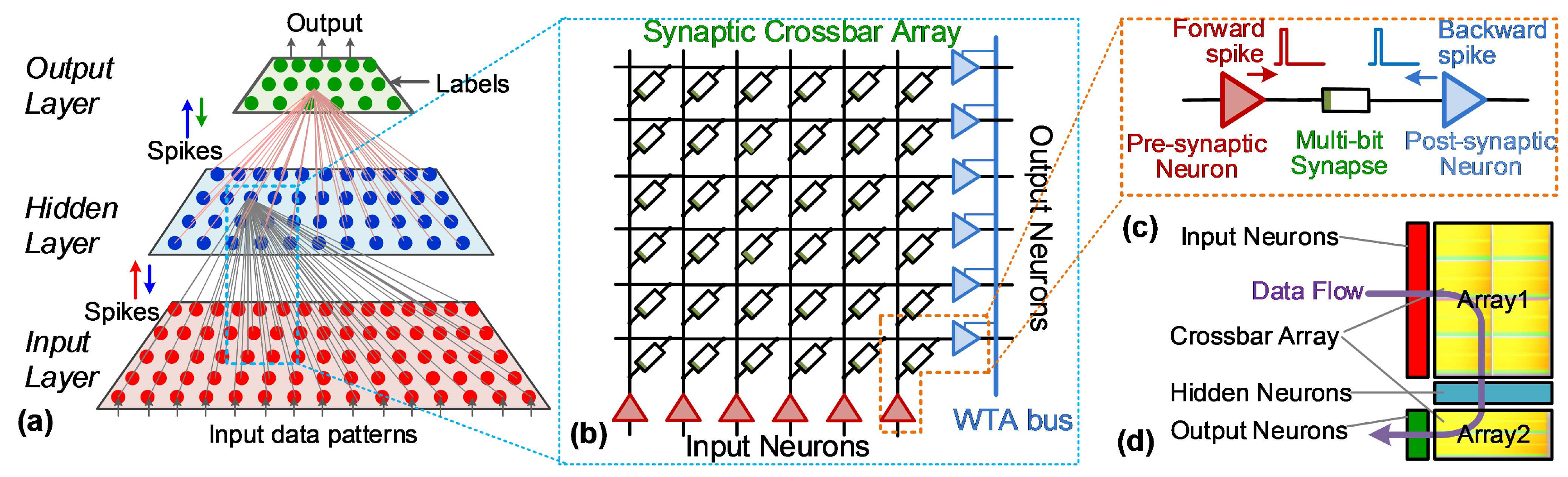}
\par\end{centering}
\caption{\label{fig:Envisioned-Neuromorphic-SoC}A neuromorphic SoC architecture:
(a) A fully-connected spiking neural network (SNN); (b) Crossbar synaptic
array and column/rows of mixed-signal CMOS neurons; (c) A synapse
between the input (pre-synaptic) and output (post-synaptic) neurons
that adjusts its weight using STDP; (d) Floorplan of 2D SNN arrays. }
\end{figure}

\IEEEPARstart{A}{} grand challenge for the semiconductor industry
is to \textquotedblleft Create a new type of computer that can proactively
interpret and learn from data, solve unfamiliar problems using what
it has learned, and operate with the energy-efficiency of the human
brain \cite{ComputingGrandChallengeReport2015}.\textquotedblright{}
Deep neural networks, or deep learning, have been remarkably successful
with a growing repertoire of problems in image and video interpretation,
speech recognition, control, and natural language processing \cite{hinton2006reducing}.
However, these implementations are compute intensive and employ high-end
servers with graphical processing units (GPUs) to train deep neural
networks. Furthermore, the new International Roadmap for Devices and
Systems (IRDS) that replaces the ITRS roadmap, looks forward to \emph{More-Moore}
and \emph{Beyond-Moore} technologies to develop radically different
data-centric computing architectures \cite{IRDS,courtland2016transistors}.
New architectures are required to transcend the device variability
and interconnect scaling bottlenecks in nano-scale CMOS, should exploit
massive parallelism, and employ in-memory computing as inspired from
biological brains. Recent progress in memristive or resistance-switching
devices (RRAM, STTRAM, phase-change memory, etc.) has spurred renewed
interest in neuromorphic computing \cite{strukov2008missing,rothenbuhler2013reconfigurable,vourkas2016memristor,kuzum2011nanoelectronic,jo2009high,gupta2009w,burr2017neuromorphic}.
Such memristive devices, integrated with standard CMOS technology,
are expected to realize low-power neuromorphic system-on-a-chip (NeuSoC)
with embedded deep learning and orders of magnitude lower power consumption
than GPUs, as illustrated in Fig. \ref{fig:Envisioned-Neuromorphic-SoC}
\cite{indiveri2013integration,strukov2006reconfigurable}. Since `analog'
memristor device technology is yet to mature while practical demonstration
in neural circuits are being pursued \cite{wu2016enabling,saxena2017towards,schuman2017survey},
we earlier proposed a low-risk and robust alternative for circuit
prototyping using a CMOS memristor emulator \cite{saxena2015memory,Saxena2017Memristor}.
In this work, we extend this CMOS memristor concept to memristive
synapse circuits that realize bio-plausible spike-timing dependent
plasticity (STDP) learning. The rest of the manuscript is organized
as follows: \textcolor{black}{Section II presents energy-estimation
of memristive NeuSoCs; Section III and IV describe the CMOS memristor
and synapse circuits. Finally Section V presents simulation results
and application in an image classification task, followed by conclusion.}

\section{Energy-efficiency of Neuromorphic SoCs}

The primary motivation of exploring memristive (or emerging NVM-based)
spiking neural network is to achieve orders of magnitude energy-efficiency
improvement over the contemporary digital architectures. This is expected
to be achieved by employing event-driven asynchronous spiking neural
networks (SNNs), with low-power circuits and ultra-low-power synaptic
(memory) devices. In an SNN, the spike shape parameters and the \emph{low-resistance
state (LRS)} resistance, $R_{LRS}$, of the memristive devices ($R_{HRS}$
is typically order(s) of magnitude higher than $R_{LRS}$) contribute
to the energy consumed in an spike event. The total energy consumption
is also decided by the sparsity, i.e. the percentage of synapses in
LRS state, spiking activity, and the power consumption in the CMOS
neurons. Assuming a rectangular spike pulse-shape of amplitude $V_{p}$
and width $T_{p}$, the current input signal is $I_{syn}=\frac{V_{p}}{R_{M}}$,
and the energy consumption for a spike driving a synapse with resistance
$R_{M}$, $R_{LRS}<R_{M}<R_{HRS}$, is given by $E_{spk}=\frac{V_{p}^{2}T_{p}}{R_{M}}<\frac{V_{p}^{2}T_{p}}{R_{LRS}}$.

The approximate SNN energy consumption for one event can be formulated
as

\begin{equation}
E_{_{SNN}}=\eta_{sp}\eta_{LRS}N_{s}E_{spk}+N_{n}P_{n}T_{p}\label{eq:E_SNN}
\end{equation}

where $\eta_{sparsity}$ is the sparsity factor (i.e. the fraction
of neurons firing on average), $\eta_{LRS}$ is fraction of synapses
in the LRS-state, $N_{s}$ is the number of synaptic connections,
$N_{n}$ is the number of neurons. $P_{neuron}$ is the neuron power
consumption; energy consumed in the peripheral circuits is ignored
to simplify the analysis. To provide a rough system-level comparison,
the AlexNet convolutional neural network for deep learning used for
the Imagenet Challenge comprised of 61M synaptic weights and 640k
neurons \cite{krizhevsky2012imagenet}. We assume that an equivalent
SNN is constructed through transfer learning \cite{diehl2015fast},
or spike-based equivalent of backpropagation algorithm \cite{neftci2016event};
the circuit architecture is essentially the same. With an estimation
based on the RRAM-compatible spiking neuron chip realized in \cite{wu2015cmos},
4-bit compound memristive synapses\cite{bill2014compound,wu2016enabling,saxena2017towards},
and $R_{LRS}$ ranging from 0.1-10$M\Omega$, the energy consumption
for processing (training or classification) of one image is shown
in Table \ref{tab:Energy-Efficiency-Estimation}. By comparing with
the contemporary advanced GPU Nvidia P4 \cite{Nvidia2016} (170 images/s/W),
a memristive architecture with $R_{LRS}=100k\Omega$ provides a meagre
$14\times$ improvement in energy-efficiency. However, the energy
consumption can be significantly reduced if the LRS resistance of
the memristive devices can be increased to high-$M\Omega$ regime,
leading to a potential $1000\times$ range performance improvement;
high LRS also helps reduce the power consumption in the opamp-based
neuron circuits \cite{wu2015cmos,saxena2009indirect}.Since there
has been less focus on realizing high-LRS devices as the multi-valued
memristive devices are still under development, circuit solutions
are desired to address this wide energy-efficiency gap. 

\begin{table}[tb]
\caption{\label{tab:Energy-Efficiency-Estimation}Energy estimation for a memristive
SNN}
\centering{}%
\begin{tabular}{|>{\raggedright}m{1in}|c|c|c|c|}
\hline 
 &  & {\scriptsize{}Low} & {\scriptsize{}Medium} & {\scriptsize{}High }\tabularnewline
\hline 
\hline 
\multirow{1}{1in}{{\scriptsize{}Spike Width}} & {\scriptsize{}$T_{p}$} & \multicolumn{3}{c|}{{\scriptsize{}100ns}}\tabularnewline
\hline 
\multirow{1}{1in}{{\scriptsize{}Spike Amplitude}} & {\scriptsize{}$V_{P}$} & \multicolumn{3}{c|}{{\scriptsize{}300mV}}\tabularnewline
\hline 
{\scriptsize{}ON State Resistance} & {\scriptsize{}$R_{LRS}$} & {\scriptsize{}$100k\Omega$} & {\scriptsize{}$1M\Omega$} & {\scriptsize{}$10M\Omega$}\tabularnewline
\hline 
{\scriptsize{}Single Spike Energy } & {\scriptsize{}$E_{spk}$} & {\scriptsize{}1.4pJ} & {\scriptsize{}140fJ} & {\scriptsize{}14fJ}\tabularnewline
\hline 
{\scriptsize{}Neuron Energy } & {\scriptsize{}$E_{N}$} & {\scriptsize{}1.56pJ} & {\scriptsize{}260fJ} & {\scriptsize{}43.3fJ}\tabularnewline
\hline 
{\scriptsize{}Neuron Sparsity} & {\scriptsize{}$\eta_{sp}$} & \multicolumn{3}{c|}{{\scriptsize{}0.6}}\tabularnewline
\hline 
{\scriptsize{}On State RRAM Ratio} & {\scriptsize{}$\eta_{LRS}$} & \multicolumn{3}{c|}{{\scriptsize{}0.5}}\tabularnewline
\hline 
{\scriptsize{}Single Event Energy} & {\scriptsize{}$E_{SNN}$} & {\scriptsize{}$422.6\mu J$} & {\scriptsize{}$42.33\mu J$} & {\scriptsize{}$4.244\mu J$}\tabularnewline
\hline 
{\scriptsize{}Images / Sec / Watt} &  & {\scriptsize{}2.4k} & {\scriptsize{}23.6k} & {\scriptsize{}235k}\tabularnewline
\hline 
\multicolumn{2}{|l|}{{\scriptsize{}Acceleration over GPU\cite{Nvidia2016}}} & {\scriptsize{}$\text{\texttimes}14$} & {\scriptsize{}$\times139$} & {\scriptsize{}$\times1.38k$}\tabularnewline
\hline 
\end{tabular}
\end{table}

\section{CMOS Memristor }

\begin{figure}[h]
\begin{centering}
\includegraphics[width=1\columnwidth]{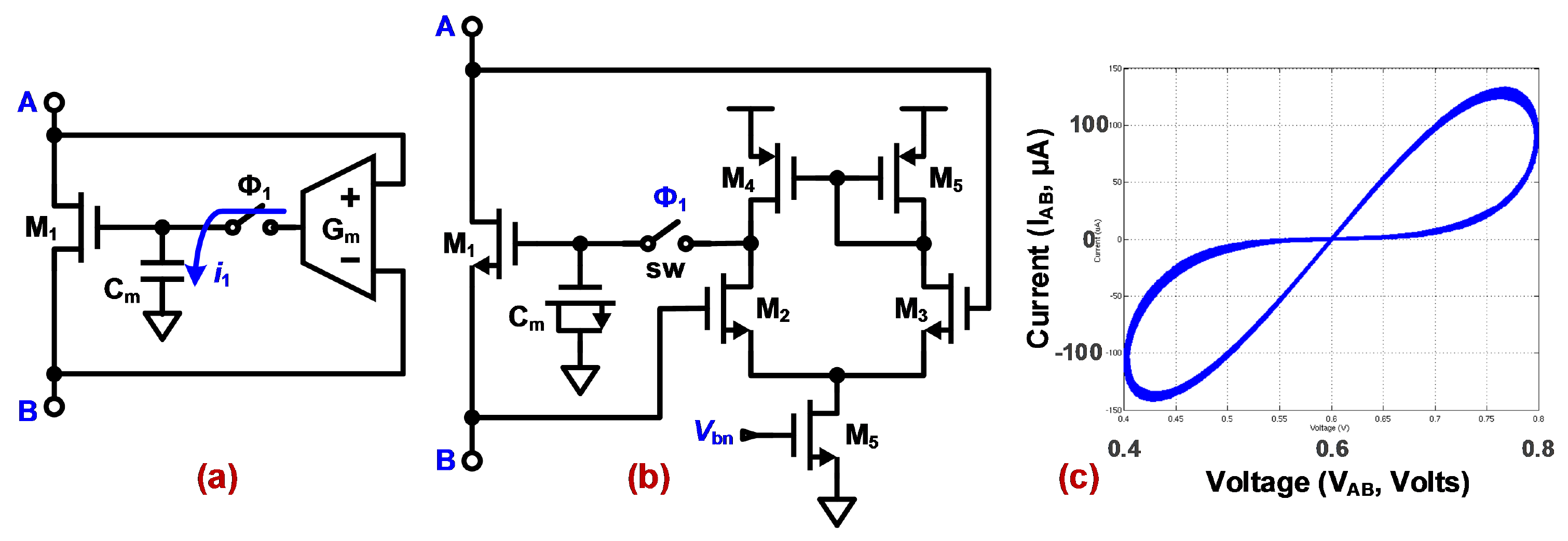}
\par\end{centering}
\caption{\label{fig:A-transistor-level-realization}A CMOS memristor circuit
with its pinched hysteresis curve \cite{saxena2015memory,Saxena2017Memristor}. }
\end{figure}

Memristor was defined as a two-terminal circuit-theoretic concept
in \cite{chua1971memristor}, and later extended to a wider class
of memristive devices \cite{chua1976memristive}. The fundamental
promise of the memristive devices lies in their `analog' memory, that
enables them to store as well as manipulate information in analog-domain.
This is harnessed in neuromorphic computing, where memristors realize
incremental synapses that learn based on STDP, a bio-inspired local
learning rule that implements spike-based expectation maximization
(SEM) algorithm \cite{indiveri2013integration,Serrano-Gotarredona2013,Nessler2013,wu2015homogeneous,saxena2017towards,neftci2016event}.
The author recently proposed a compact CMOS memristor (emulator) circuit
\cite{saxena2015memory,Saxena2017Memristor}. The fundamental concept
is illustrated in Fig. \ref{fig:A-transistor-level-realization} (a\&b),
where an n-channel MOSFET (NMOS) $M_{1}$ implements a floating variable
resistance between terminals A and B. The transconductor $G_{m}$
senses the voltage across the two terminals, produces a small-signal
current which is integrated as charge on capacitor $C_{m}$. When
the strobe $\Phi_{1}$ is low, the capacitor $C_{m}$ is disconnected
from the transconductor and holds the stored charge; thus realizing
a dynamic analog memory. 

Here, the voltage on the capacitor, $V_{G}$, controls the gate of
$M_{1}$ and is thus the \textquoteleft state,\textquoteright{} $x$,
of the synapse. The switch ($SW$) prevents $G_{m}$'s output from
leaking the state $(x\equiv V_{G}$) on capacitor $C_{m}$ when no
inputs are applied. Assuming that $M_{1}$ is in triode, the dynamics
of the memristor circuit are approximated as

\begin{equation}
I\approx\beta(V_{GS}-V_{THN})\cdot V_{AB}=G(x)\cdot V_{AB}\label{eq:mem3-1}
\end{equation}
\begin{equation}
\dot{x}=\frac{G_{m}(V_{AB})}{C_{m}}\cdot V_{AB}\equiv f(V_{AB},t)\label{eq:mem4-1}
\end{equation}

where $V_{GS}$ and $V_{THN}$ are the gate-to-source and threshold
voltages; $\beta=KP_{n}\frac{W}{L}$, $KP_{n}$ is the transconductance
parameter, $W/L$ is the sizing for $M_{1}$. In order to force $M_{1}$
in triode for large drain-source voltage swings, a zero- or low-threshold
voltage (ZVT/LVT) transistor is employed \cite{saxena2015memory,Saxena2017Memristor}.
The simulated current-voltage characteristics for the memristor circuit,
seen in Fig. \ref{fig:A-transistor-level-realization} (c), confirms
the pinched hysteresis signature of an ideal memristor. 

Contemporary memristive devices exhibit several limitations; they
exhibit stochastic switching and variability in resistance states,
depending upon the initial \textquoteleft forming\textquoteright{}
step \cite{waser2009redox,yu2011stochastic,ielmini2015resistive}.
Further, it is challenging to realize stable multi-valued weights
with filamentary devices \cite{liu2013analog,saxena2017towards};
Oxide-switching devices have exhibited \textasciitilde{}9 states and
their performance \emph{in-situ} a circuit is being investigated \cite{beckmann2016nanoscale}.
A greater impediment for realizing NeuSoCs is the lower LRS resistance
observed in memristive devices ($100\Omega-10k\Omega$)\cite{beckmann2016nanoscale},
which leads to energy inefficiency as detailed earlier in Section
II. Thus, it is desirable to realize CMOS based memristive synapses
for enabling system-level exploration while the memristive devices
mature in research. 

\section{Memristive Synapse Circuit}

\begin{figure}[h]
\begin{centering}
\includegraphics[width=1\columnwidth]{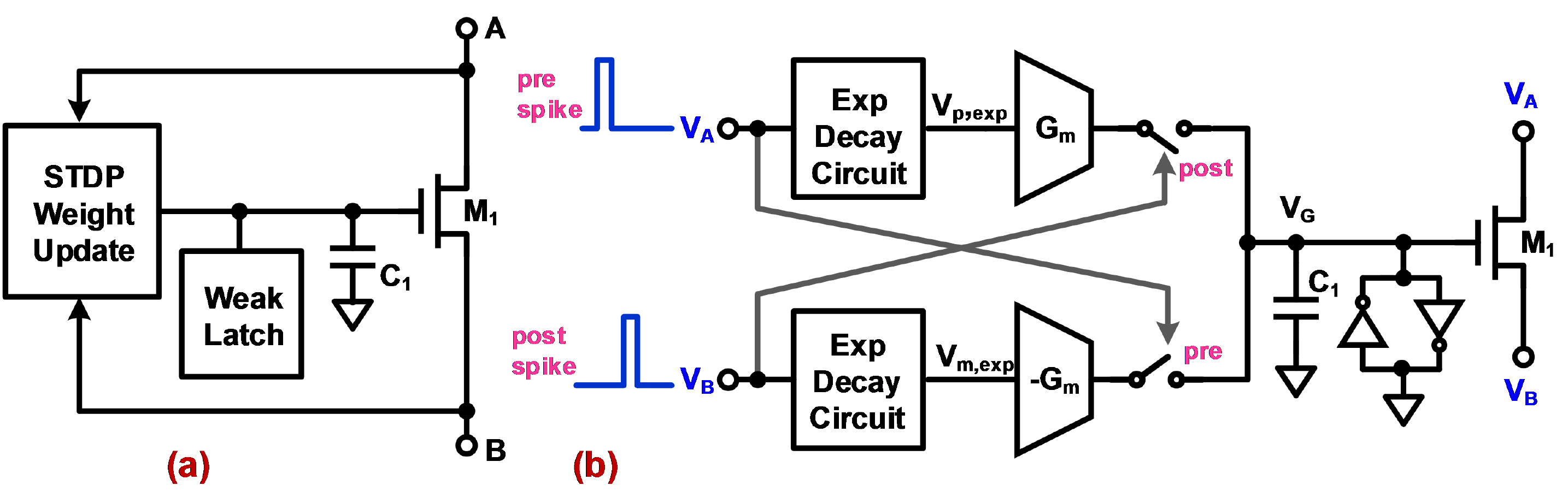}
\par\end{centering}
\caption{\label{fig:An-STDP-compatible-synapse}Memristive STDP synapse concept:
(a) the synapse with a weight update circuit to control its state,
(b) schematic showing an implementation of the synapse \cite{saxena2015memory}.}
\end{figure}

\begin{figure}[h]
\begin{centering}
\includegraphics[width=1\columnwidth]{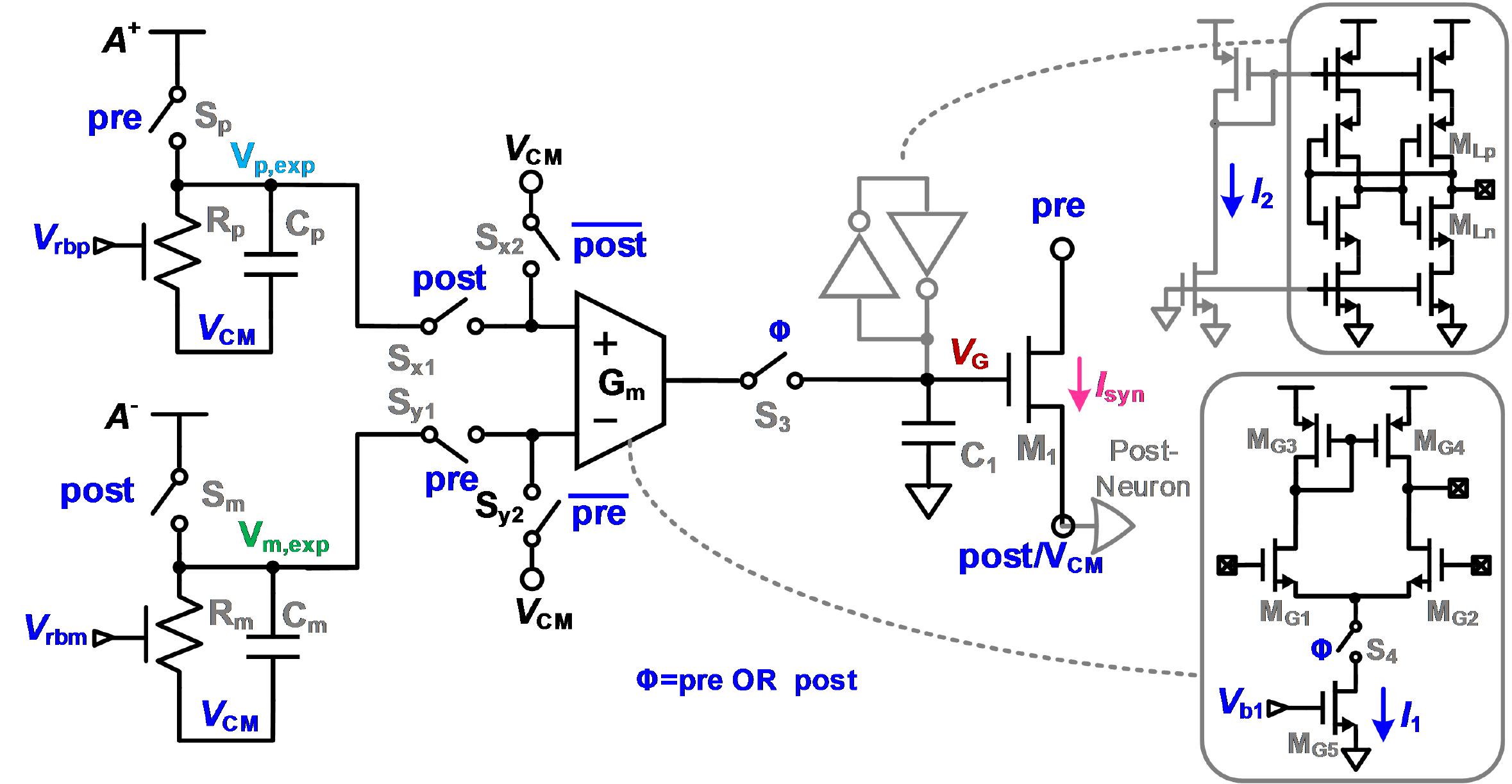}
\par\end{centering}
\caption{\label{fig:A-preferred-embodiment}A CMOS implementation of the memristive
STDP synapse with long-term bistability.}
\end{figure}

Memristive spiking circuits typically use analog spikes with rectangular
positive pulse with a negative exponential tail \cite{wu2015cmos};
however, representation of spikes with digital pulses is highly desirable
for large-scale NeuSoC implementation. Further, an accelerated neural
dynamics with moderate speed (few \emph{MHz}'s) is preferred over
biological time-scales (sub-\emph{kHz}) for optimizing CMOS circuit
area and energy consumption \cite{schemmel2010wafer}. \textcolor{black}{Current-output
type bio-mimetic synapses are pervasive in literature (\cite{liu2015event}
and references therein), where subthreshold analog techniques were
used to mimic synaptic ion-channel dynamics. Most recently, \cite{Cruz-Albrecht2012}
reported a pair-wise STDP synapse with short-term retention, and \cite{arthur2006learning}
combined subthreshold circuits with a latch.}\textcolor{red}{{} }In
contrast, we have proposed \emph{memristive} STDP-learning synapse
concept shown in Figure \ref{fig:An-STDP-compatible-synapse} and
was previously disclosed by us in \cite{saxena2015memory}. In this
work, we expand on the previous disclosure, and present circuits and
system-level details. The circuit employs the trace decay method for
emulating STDP as used in the event-driven simulators for computational
neuroscience \cite{goodman2009brian,Cruz-Albrecht2012,arthur2006learning}.\textcolor{black}{{}
}\textcolor{red}{} The STDP weight update block converts the relative
timing between \emph{pre} and \emph{post} spikes ($\Delta t=t_{post}-t_{pre}$)
into change in $V_{G}$, and thus the synaptic weight. Figure \ref{fig:An-STDP-compatible-synapse}
(b) shows schematics for the synapse, and one of several possible
transistor-level implementations is shown in Fig. \ref{fig:A-preferred-embodiment}.
Here, the input \emph{pre} and \emph{post} pulses are converted into
voltage traces $V_{p,exp}$ and $V_{m,exp}$ respectively, using the
two Exponential Decay Circuit (EDC) blocks. The outputs of the EDCs
are translated to a current using the shared $G_{m}$, which are then
integrated on $C_{1}$. The exponential trace is implemented using
active resistors, $R_{p,m}$, with time-constants $\tau_{p,m}=R_{p,m}C_{p,m}$
which can be independently tuned. 

\begin{figure}[h]
\begin{centering}
\includegraphics[width=1\columnwidth]{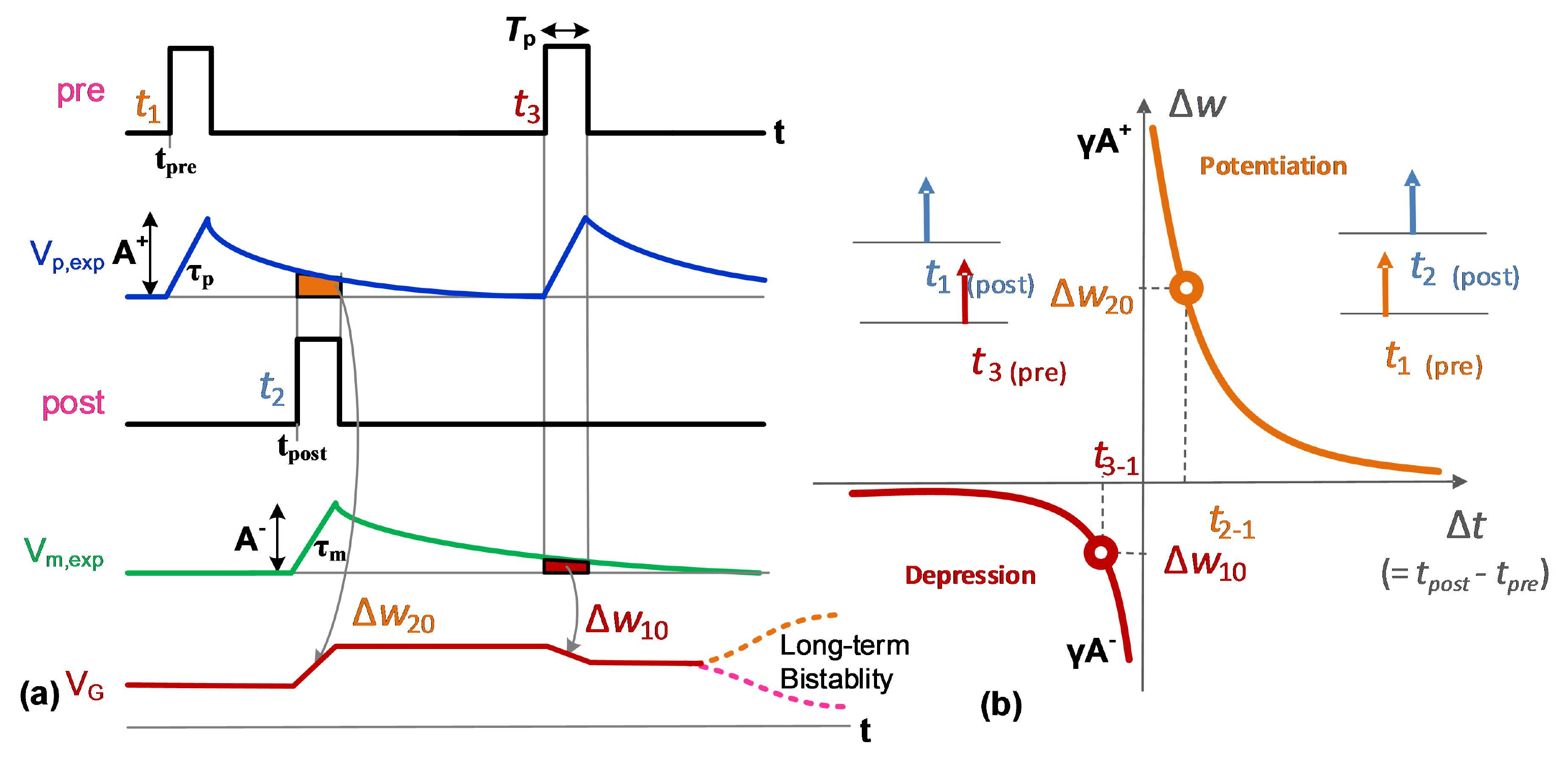}
\par\end{centering}
\caption{\label{fig:Characteristic-timing-diagram}Characteristic timing diagram
and waveforms for the pair-wise bistable STDP synapse circuit with
digital spikes, (b) STDP learning function implemented in the synapse.}
\end{figure}

Figure \ref{fig:Characteristic-timing-diagram} (a) illustrates the
synapse operation and the resulting pair-wise additive STDP learning
function is shown in Figure \ref{fig:Characteristic-timing-diagram}
(b). Here, the (pre)-synaptic spike arrives earlier than the (post)-synaptic
spike. EDC output, $V_{p,exp}$, is then sampled by the \emph{post}
spike. This sampled voltage $V_{p,exp}(\Delta t)$ leads to an increase
in the voltage $V_{G}$ (i.e. the state of the synapse) and increase
in synaptic weight/conductance ($w\equiv G(\text{V}_{G})$); synapse
undergoes \emph{short-term potentiat}ion. Similarly, in the second
case, the \emph{post} spike arrives earlier than \emph{pre} which
in turn reduces the synapse state $V_{G}$; the synapse undergoes
\emph{short-term depression}. The references $A^{+}$ and $A^{-}$
determine the maximum synaptic potentiation/depression as $\Delta w_{p,m}\approx\pm\frac{\beta G_{m}}{C_{1}}A^{\pm}T_{p}=\pm\gamma A^{\pm}$.

Even though the dynamic STDP synapses provide analog states, they
can only realize short-term potentiation/depression as the capacitor
memory leaks away in few milli-seconds. However in a NeuSoC, the final
weights after training must be persistent and amenable for read-out/in.
This is realized by employing\emph{ long-term bistability} in synapses
where after short-term STDP learning, the weights are quantized to
either a high or low binary conductance state. As shown in Figs. \ref{fig:An-STDP-compatible-synapse}\&\ref{fig:A-preferred-embodiment},
a weak latch is connected to $V_{G}$. This slow resolving subthreshold
latch is designed for very large regeneration time-constants ($\tau_{w}=$1-5ms)
such that it doesn\textquoteright t interfere with the short-term
STDP learning. However, once the STDP pulses are no longer present,
the weak bistable latch slowly steers the state of the synapse to
either a large voltage (LRS) or to a low-voltage (HRS) long-term states,
which can easily be read-out. 
\begin{table}[h]
\caption{\label{tab:Design-parameters-for}Design parameters for the synapse
circuit implemented in 130nm CMOS with $L_{min}=0.12\mu m$ and $V_{DD}=1.2V$. }
\centering{}%
\begin{tabular}{|c|c|c||c|c|}
\hline 
\textcolor{black}{\scriptsize{}Device} & \textcolor{black}{\scriptsize{}Type} & \textcolor{black}{\scriptsize{}W/L, $\mu m/\mu m$} & \textcolor{black}{\scriptsize{}Parameter} & \textcolor{black}{\scriptsize{}Value}\tabularnewline
\hline 
\hline 
\textcolor{black}{\scriptsize{}$M_{G1,2}$} & \textcolor{black}{\scriptsize{}NMOS} & \textcolor{black}{\scriptsize{}$1.2/0.12$} & \textcolor{black}{\scriptsize{}$C_{1}$} & \textcolor{black}{\scriptsize{}500fF}\tabularnewline
\hline 
\textcolor{black}{\scriptsize{}$M_{G3,4}$} & \textcolor{black}{\scriptsize{}PMOS} & \textcolor{black}{\scriptsize{}$1.2/0.12$} & \textcolor{black}{\scriptsize{}$C_{p,m}$} & \textcolor{black}{\scriptsize{}100fF}\tabularnewline
\hline 
\textcolor{black}{\scriptsize{}$M_{Ln}$} & \textcolor{black}{\scriptsize{}NMOS} & \textcolor{black}{\scriptsize{}$0.16/0.12$} & \textcolor{black}{\scriptsize{}$I_{1}$} & \textcolor{black}{\scriptsize{}10nA}\tabularnewline
\hline 
\textcolor{black}{\scriptsize{}$M_{Lp}$} & \textcolor{black}{\scriptsize{}PMOS} & \textcolor{black}{\scriptsize{}$0.16/0.12$} & \textcolor{black}{\scriptsize{}$I_{2}$} & \textcolor{black}{\scriptsize{}\textasciitilde{}100pA}\tabularnewline
\hline 
\textcolor{black}{\scriptsize{}$S_{p,m}$} & \textcolor{black}{\scriptsize{}PMOS} & \textcolor{black}{\scriptsize{}$1.2/0.12$} & \textcolor{black}{\scriptsize{}$I_{lk_{p,m}}$} & \textcolor{black}{\scriptsize{}\textasciitilde{}90pA}\tabularnewline
\hline 
\textcolor{black}{\scriptsize{}$S_{x,y,3,4}$} & \textcolor{black}{\scriptsize{}NMOS} & \textcolor{black}{\scriptsize{}$1.2/0.12$} & \textcolor{black}{\scriptsize{}$I_{lk_{OTA}}$} & \textcolor{black}{\scriptsize{}\textasciitilde{}110pA}\tabularnewline
\hline 
\textcolor{black}{\scriptsize{}$M_{Rp,m}$} & \textcolor{black}{\scriptsize{}NMOS} & \textcolor{black}{\scriptsize{}$0.24/0.12$} &  & \tabularnewline
\hline 
\hline 
\multirow{2}{*}{\textcolor{black}{\scriptsize{}$M_{1}$}} & \multirow{2}{*}{\textcolor{black}{\scriptsize{}LVT NMOS }} & \multicolumn{1}{c|}{\multirow{2}{*}{\textcolor{black}{\scriptsize{}$0.16/3.6$}}} & \textcolor{black}{\scriptsize{}LRS} & \textcolor{black}{\scriptsize{}$0.4M\Omega$}\tabularnewline
\cline{4-5} 
 &  &  & \textcolor{black}{\scriptsize{}HRS} & \textcolor{black}{\scriptsize{}$16M\Omega$}\tabularnewline
\hline 
\end{tabular}
\end{table}

The synapse circuit in Fig. \ref{fig:A-preferred-embodiment} is implemented
in a 130nm CMOS technology with supply $V_{DD}=1.2V$. The transistor
sizing and parameter values used in this circuit are listed in Table
\ref{tab:Design-parameters-for}. The memristive synapse, for the
given sizing for $M_{1}$, realizes LRS and HRS resistances of $0.4M\Omega$
and $16M\Omega$ respectively, providing significant improvement over
contemporary memristive devices. As detailed in \cite{wu2015cmos,wu2015homogeneous},
the traditional subthreshold neuron designs are not suitable for driving
memristive load. The opamp-based integrate-and-fire neurons with winner-take-all
STDP learning interface from author's prior work in \cite{wu2015homogeneous}
can directly be adapted to interface with the presented synapses;
higher LRS resistance will further help simply opamp design. 

\section{Simulation Results}

\begin{figure}[h]
\begin{centering}
\includegraphics[width=0.5\columnwidth]{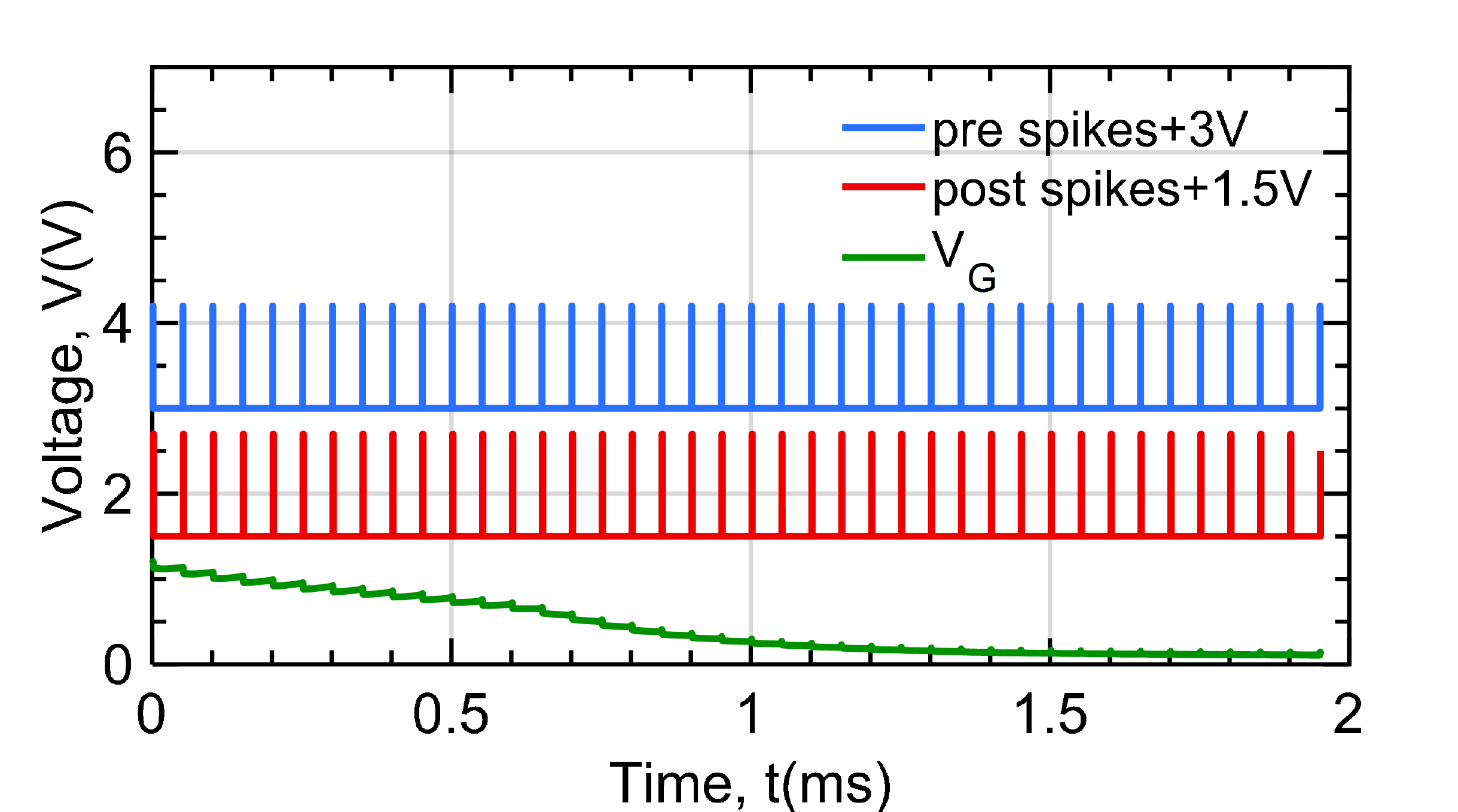}\includegraphics[width=0.5\columnwidth]{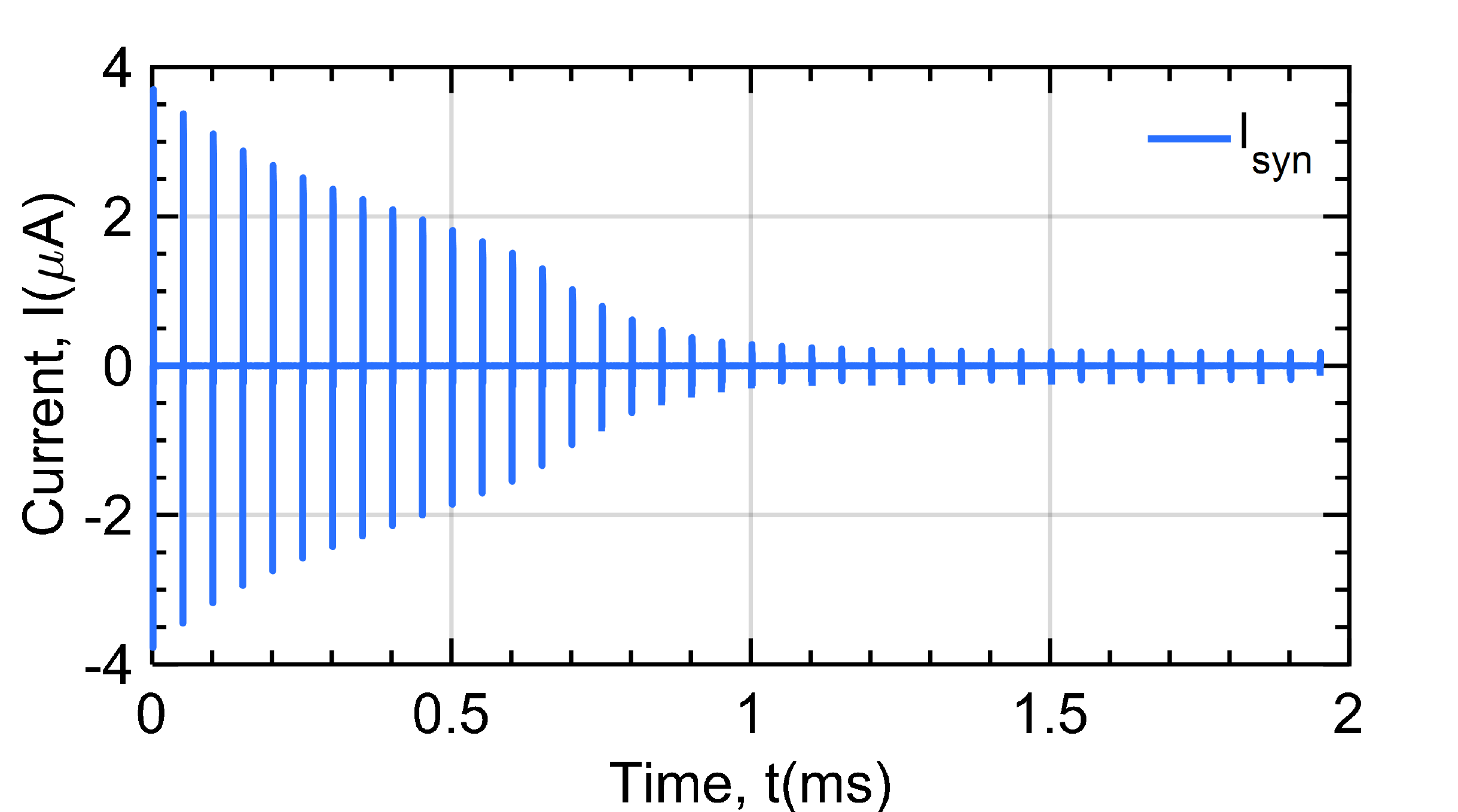}
\par\end{centering}
\caption{\label{fig:Transient-simulation-for}Transient simulation for a single
synapse for $\Delta t=t_{post}-t_{pre}=-1\mu s$: (left) the state,
$V_{G}$, is incrementally decreased, resulting in (right) corresponding
change in the conductance (synaptic current, $I_{syn}$) of the synapse.}
\end{figure}

In this design, the total standby current drawn from $V_{DD}$ is
$\simeq490pA$, while 10.4nA is drawn during the pre/post spike event.
This results in a static power consumption of 588pW and dynamic energy
consumption of 91.24fJ/spike (for $V_{pre}-V_{post}=600mV$) in the
LRS state. This circuit can be easily modified to different specifications
and further optimized for energy-efficiency, area and speed. Figure
\ref{fig:Transient-simulation-for} shows transient simulation for
a single synapse; pre and post pulses are applied with $\Delta t=t_{post}-t_{pre}=-1\mu s$,
spaced by $50\mu s$ and the state voltage $V_{G}$ and synaptic current
between pre and post terminals are displayed. We can observe that
the weight undergoes monotonic decrease due to pair-wise STDP updates
with a corresponding decrease in synapse weight/conductance, $w$,
and thus the synaptic current, $I_{syn}=w(V_{pre}-V_{post})$.

\begin{figure}[h]
\begin{centering}
\includegraphics[width=0.5\columnwidth]{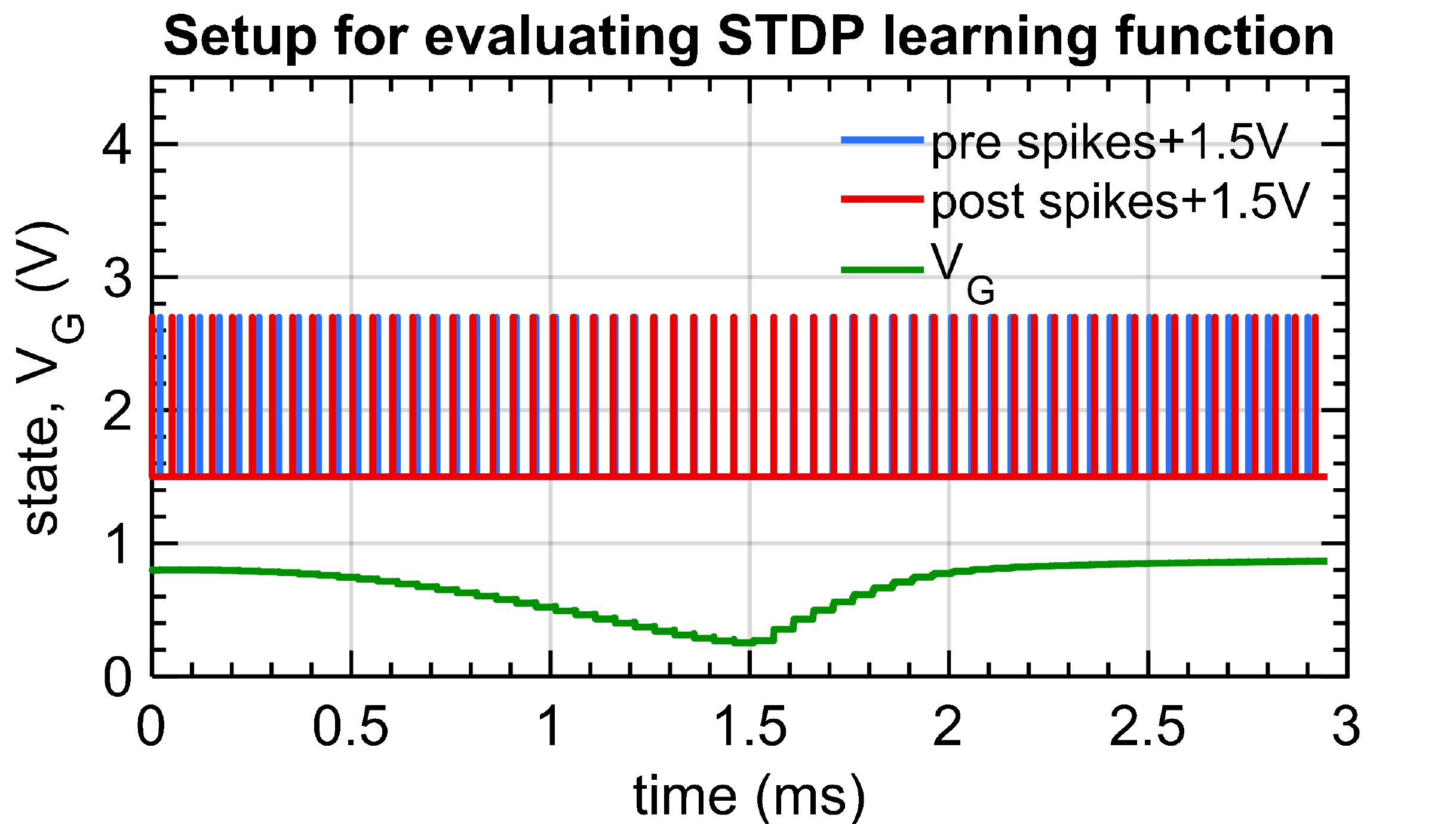}\includegraphics[width=0.5\columnwidth]{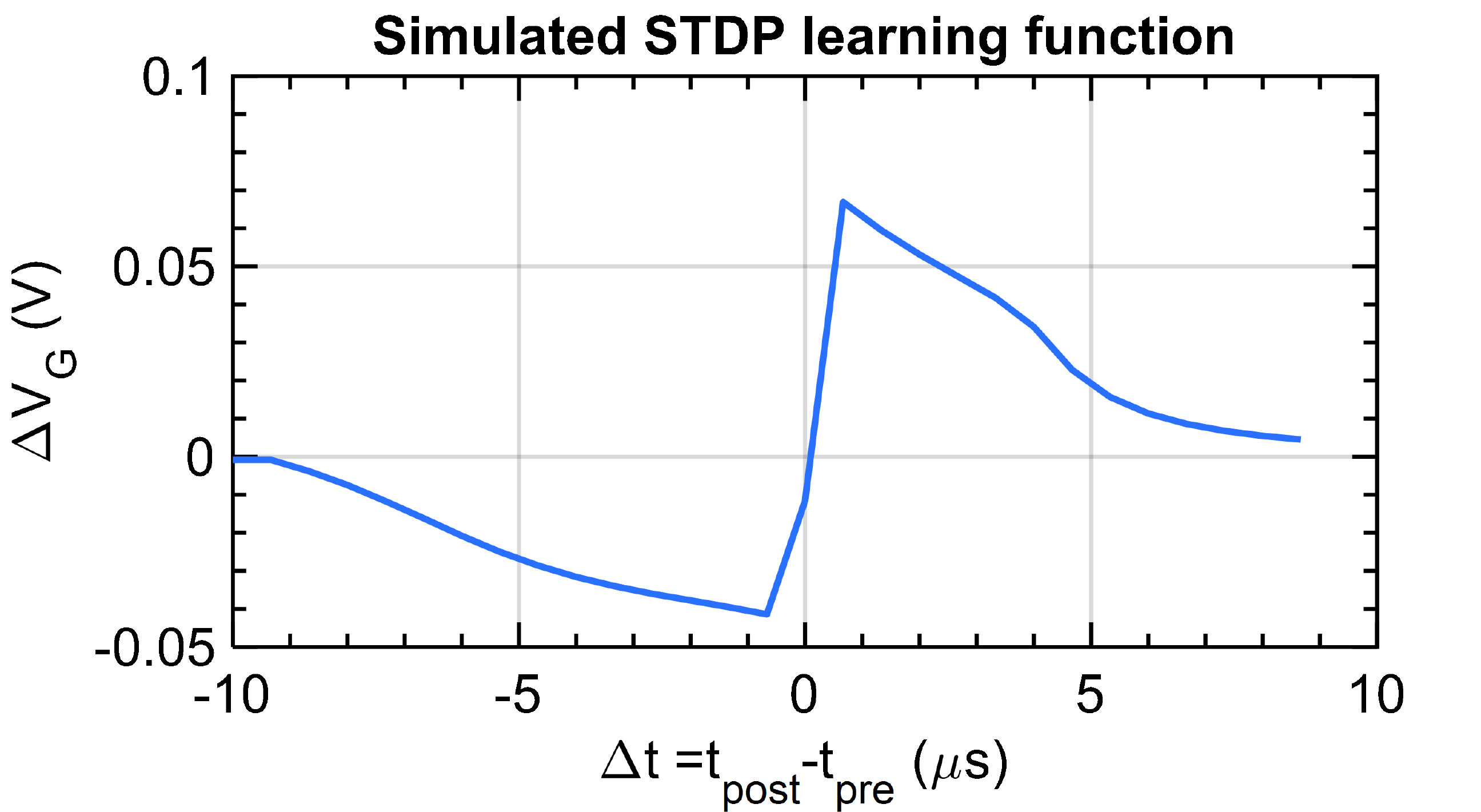}\caption{\label{fig:Simulated-STDP-learning}(left) Simulation for determining
the STDP learning function, (right) simulated STDP learning function
for the synapse circuit parameters in Table \ref{tab:Design-parameters-for}.}
\par\end{centering}
\end{figure}

Next, a transient simulation shown in Fig. \ref{fig:Simulated-STDP-learning}
is constructed to determine the STDP learning function for the synapse
circuit. Here, pre and post spikes are applied with progressively
changing $\Delta t$ from $-10\mu s$ to $10\mu s$ with spacing of
$50\mu s$ to allow the transients to completely decay. This results
in approximate double-exponential learning function characteristic
of pair-wise STDP function seen in Fig. \ref{fig:Characteristic-timing-diagram}(b).

\begin{figure}[h]
\begin{centering}
\includegraphics[width=0.5\columnwidth]{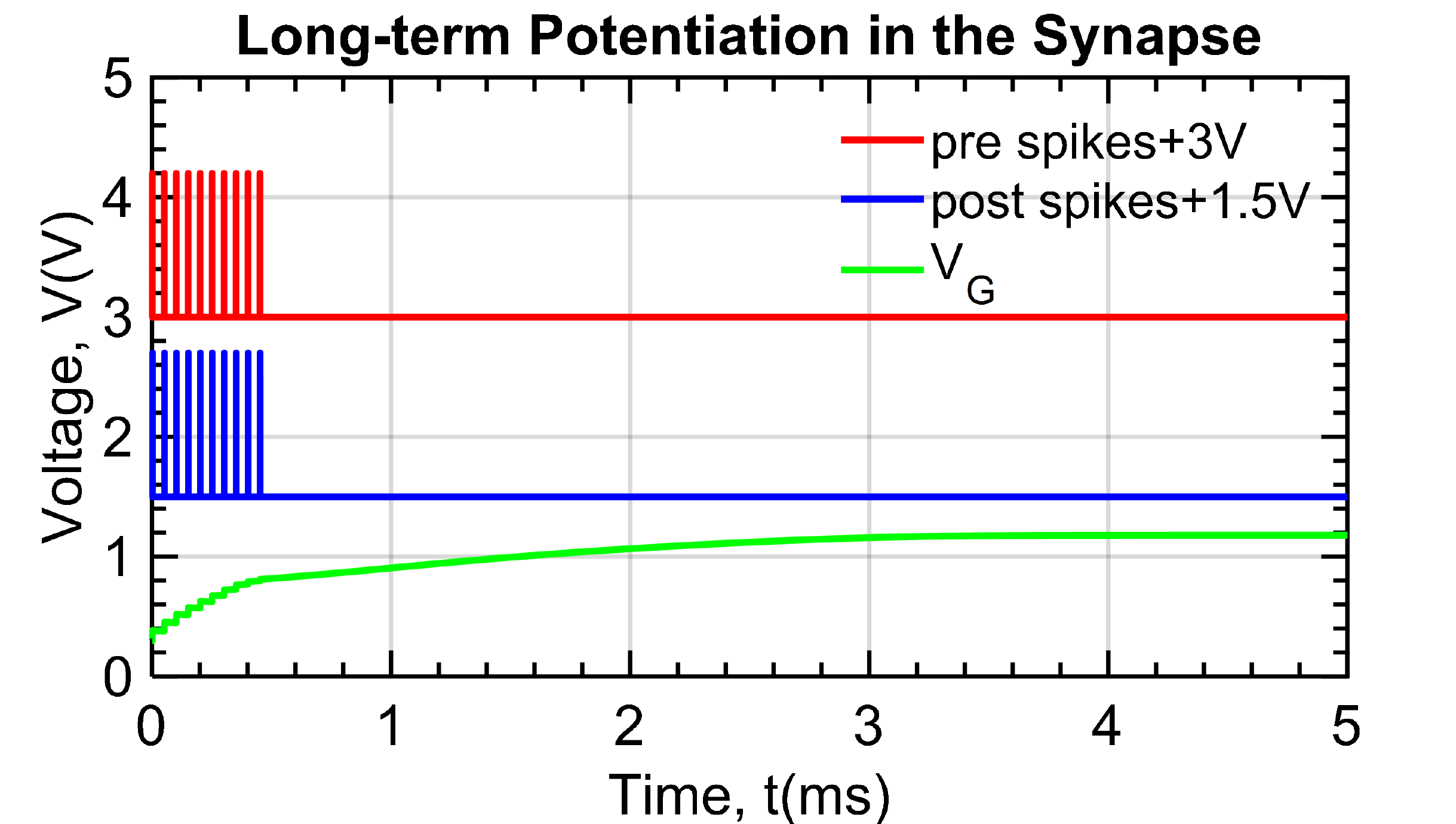}\includegraphics[width=0.5\columnwidth]{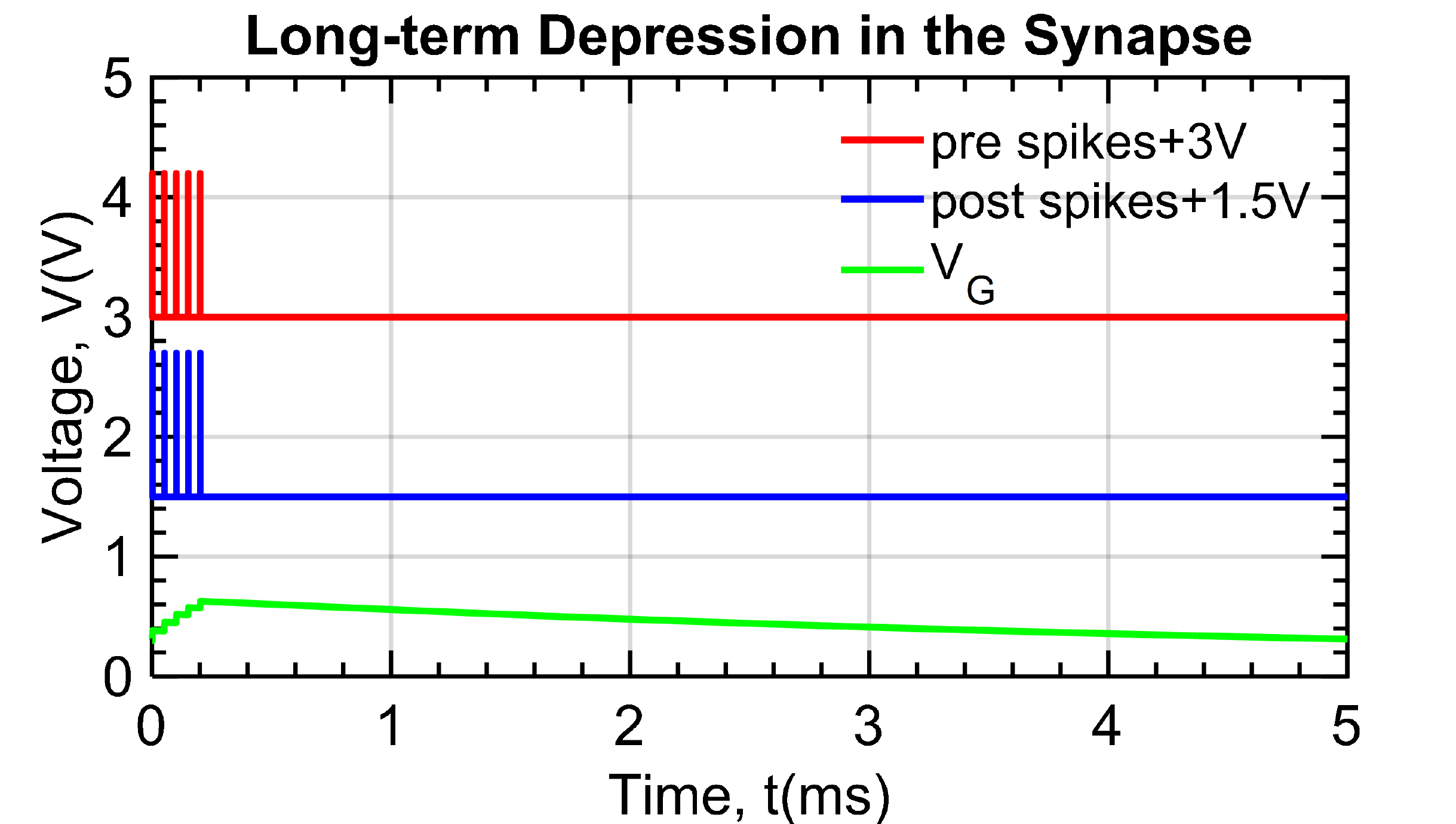}
\par\end{centering}
\caption{\label{fig:Transient-simulations-demonstrat}Transient simulations
demonstrating long-term bistability affected by the weak latch with
$\Delta t=t_{post}-t_{pre}=1\mu s$: (a) long-term potentiation, (b)
long-term depression. }

\end{figure}

Long-term bistability is demonstrated through simulations in Fig.
\ref{fig:Transient-simulations-demonstrat} where spikes are applied
such that the weight crosses the latch's threshold point, $V_{w,thr}\approx0.6V$
in Fig. \ref{fig:Transient-simulations-demonstrat}(left) and is below
threshold in Fig. \ref{fig:Transient-simulations-demonstrat} (right).
The weak latch is biased in subthreshold and has a regenerative time-constant
of $\tau_{w}\approx2ms$; the latch slowly resolves the synaptic state
to logic high (LRS) or low (LRS). 

\begin{figure}[h]
\begin{centering}
\includegraphics[width=0.8\columnwidth]{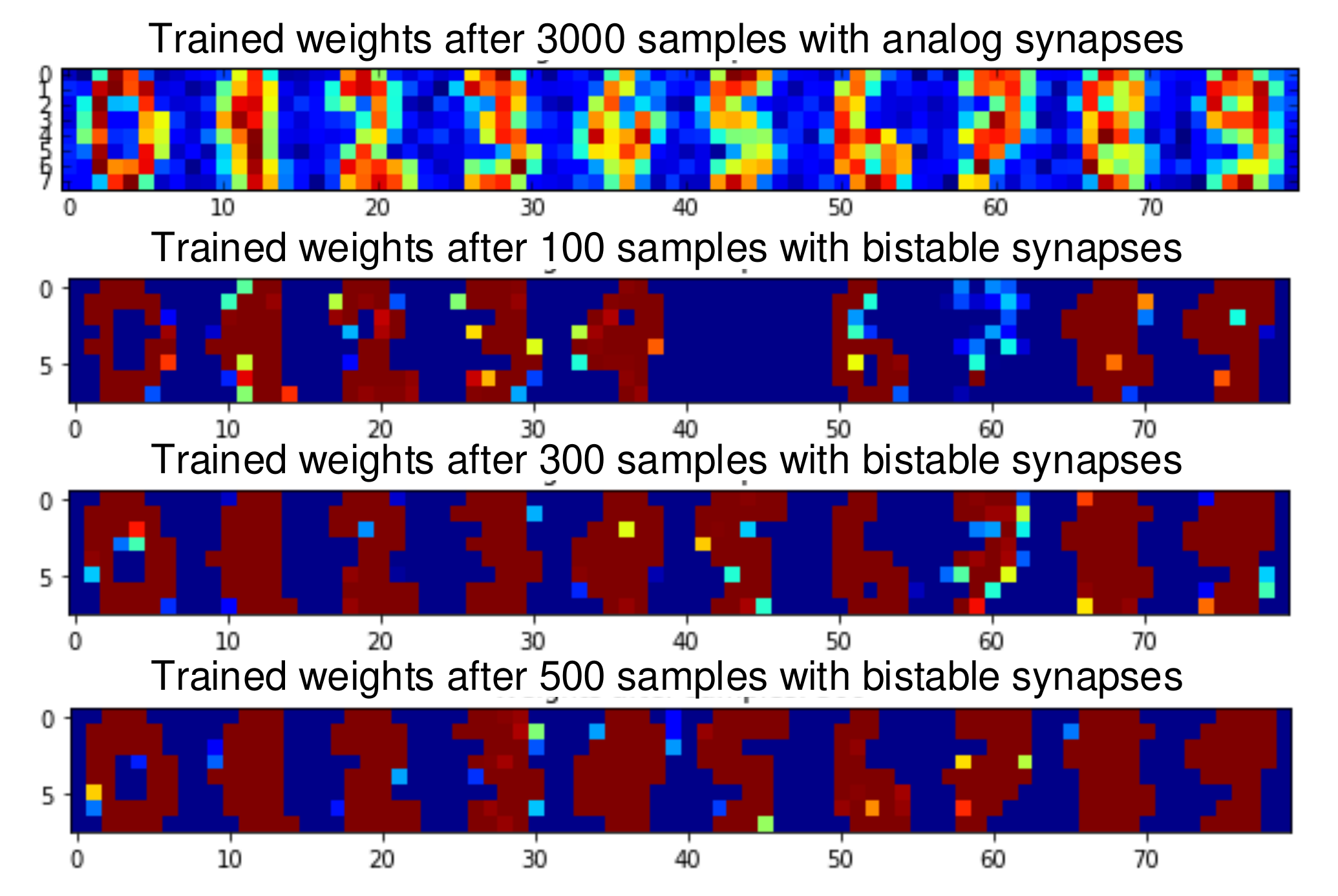}\caption{\label{fig:SNN-example-for}SNN for handwritten digits classification.
Synaptic weight evolution rearranged as 8\texttimes 8 bitmap for analog
and the proposed bistable synapses. }
\par\end{centering}
\end{figure}

An SNN, similar to \cite{wu2015homogeneous}, was setup using the
bistable memristive synapse and winner-take-all neuron macro-models
using Brian2 libraries in Python \cite{goodman2009brian}. The UCI
handwritten digits dataset (3,823 training and 1,797 test $8\times8$
bitmap images \cite{bache2013uci}) was used to train the fully-connected
SNN with 64 input and 10 output neurons, 640 synapses, and with a
teacher signal enforcing the output labels. Fig. \ref{fig:SNN-example-for}
shows the learned weights for each output neuron. For analog synapses
the test accuracy was 83\% for all 10 digits (96\% for 4 digits);
the bistable synapses achieve accuracy of \textasciitilde{}74\% for
10 digits due to binary quantization during training. Care must be
taken to ensure that $\tau_{w}$ is much larger than the time for
which input samples are presented ($50\mu s$) to avoid\emph{ }catastrophic
forgetting. In this experiment, the bistable SNN was trained for 500
images as a large number of weights, $w_{ijk}$, start approaching
$w_{max}=1$, resulting in loss of classification accuracy. Further,
$w_{min}\apprge0.01$ must be used as otherwise there is a chance
of all the weights getting quantized to 0, and the neurons will never
fire. 

\section{Conclusion\label{sec:Conclusion}}

A compact analog memristive STDP synapse circuit, with long-term binary
retention and high LRS resistance, is introduced and designed in standard
CMOS and analytical as well as simulation results are presented. The
circuit is used to realize image classification application and the
challenges are discussed. In summary, the synapse provides an efficient
circuit solution for NeuSoC architecture exploration, while memristive
devices on CMOS platforms reach maturity. 

\section*{}

\bibliographystyle{IEEEtran}
\bibliography{Memristors}

\end{document}